\documentclass[conference]{IEEEtran}
\IEEEoverridecommandlockouts
\usepackage[nolist]{acronym}
\usepackage{tikz}
\usepackage{pgfplots}
\usepackage{array}
\usepackage{algorithm,algorithmicx}
\usepackage{algpseudocode}
\usepackage{amsmath}
\usepackage{cleveref}
\usepackage[numbers,sort&compress]{natbib}
\usepackage{multirow}
\usepackage{caption}
\usepackage{subcaption}
\usepackage{amsmath}
\def\BibTeX{{\rm B\kern-.05em{\sc i\kern-.025em b}\kern-.08em
    T\kern-.1667em\lower.7ex\hbox{E}\kern-.125emX}}
\begin{document}

\title{Train Localization During GNSS Outages: A
Minimalist Approach Using Track Geometry And
IMU Sensor Data\\
\thanks{This work has been funded by Trafikverket (The Swedish Transport Administration),  through Excellence Area 6, TRV 2024/2731.}

}

\author{\IEEEauthorblockN{Wendi Löffler}
\IEEEauthorblockA{\textit{Information Science and Engineering} \\
\textit{KTH Royal Institute of Technology}\\
Stockholm, Sweden \\
loeffler@kth.se}
\and
\IEEEauthorblockN{Mats Bengtsson}
\IEEEauthorblockA{\textit{Information Science and Engineering} \\
\textit{KTH Royal Institute of Technology}\\
Stockholm, Sweden \\
matben@kth.se}
}

\maketitle

\begin{abstract}
Train localization during \ac{GNSS} outages presents challenges for ensuring fail-safe and accurate positioning in railway networks. This paper proposes a minimalist approach exploiting track geometry and \ac{IMU} sensor data. By integrating a discrete track map as a Look-Up Table (LUT) into a Particle Filter (PF) based solution, accurate train positioning is achieved with only an \ac{IMU} sensor and track map data. The approach is tested on an open railway positioning data set, showing that accurate positioning (absolute errors below 10\,m) can be maintained during \ac{GNSS} outages up to 30\,s in the given data. We simulate outages on different track segments and show that accurate positioning is reached during track curves and curvy railway lines. The approach can be used as a redundant complement to established positioning solutions to increase the position estimate's reliability and robustness. 
\end{abstract}


\begin{acronym}
 \acro{ETCS}{European Train Control System}
 \acro{GNSS}{Global Navigation Satellite Systems}
 \acro{IMU}{Inertial Measurement Unit}
 \acro{SLAM}{Simultaneous localization and mapping}
 \acro{PF}{Particle Filter}
 \acro{EKF}{Extended Kalman Filter}
 \acro{LUT}{Look-Up Table}
 \acro{ENU}{East North Up}
 \acro{DTM}{Discrete Track Map}
 \acro{ZVU}{Zero Velocity Update}
 \acro{KF}{Kalman Filter}
 \acro{CDF}{Cumulative Density Function}
 \acro{EKFMM}{Extended Kalman Filter with Map Matching}
 \acro{CTRA}{Constant Turn Rate and Acceleration}
\end{acronym}

\section{Introduction}
\label{sec:introduction}
Determination of train positions within the railway network must be fail-safe and of high accuracy. Inaccurate train position data can lead to delays in operation \cite{hamid2020impact}. In systems operating today, train position is determined using trackside equipment such as axle counters and balises. These solutions are vulnerable to vandalism and weather influence; hence, to reduce maintenance costs on the trackside, train-borne solutions that are not dependent on trackside equipment have been a topic of interest in research for some years. 

A widely used approach in tracking problems is fusion of Global Navigation Satellite Systems (\ac{GNSS}) and Inertial Measurement Units (\ac{IMU}) \cite{gustafsson2010statistical}. The challenges consist of adjusting the method to the high safety standards according to EN50126 \cite{en50126} and the specifics of the railway environment.
Several \ac{KF} based solutions have been proposed using \ac{GNSS}, \ac{IMU}, and odometer data \cite{reimer2016ins,otegui2018evaluation}.

\begin{figure}[h!]
    \centering
    \includegraphics[width=0.7\linewidth]{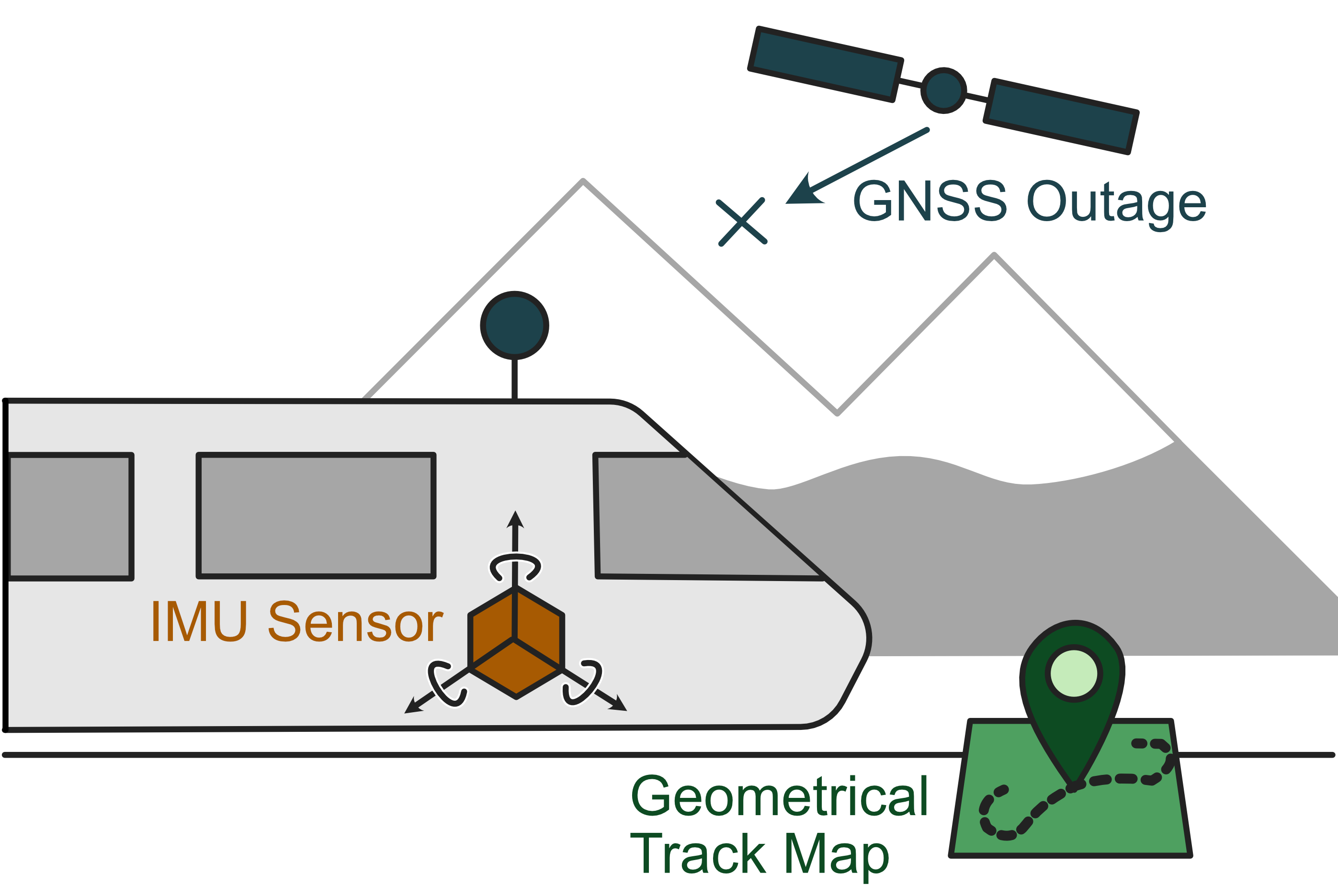}
    \caption{Considered set up in this paper. Our proposed solution exploits an IMU and a track map.}
    \label{fig:introductionscheme}
\end{figure}

Unlike other transport modes such as aviation and automotive, train motion is restricted to the tracks; a constraint that can be used to improve the accuracy of the position estimate and different methods have been proposed fusing \ac{GNSS}/\ac{IMU}, and a track map \cite{heirich2016bayesian,liu2016track, liu2023robust, winter2018increasing}. Geometric track maps can be used for train positioning in different ways: in \cite{heirich2013railslam, loffler2022using, loffler2022evaluating}, the authors use linear interpolation between discrete coordinates, whereas in \cite{hasberg2011simultaneous}, spline interpolation is used. On the other hand, \cite{winter2019generating} propose a formulation of the track map based on the geometrical elements in track construction, promising a more precise representation. However, \cite{andersson2023compact} shows that linear interpolation between discrete coordinates with feasible distance offers sufficient accuracy in most applications.

Given the high safety requirements, \ac{GNSS} cannot provide the required availability and accuracy in the complex propagation environments, which are encountered in railway settings \cite{spinsante2020hybridized,zhu2018gnss}. Tunnels, urban areas, heavily wooded districts, or deep valleys cause \ac{GNSS} outages. Polar regions, as relevant in Sweden, cause a decrease in \ac{GNSS} accuracy. In  \cite{d2023innovative} an extensive study is performed using positioning data from 80.000\,km of rails and observe \ac{GNSS} outages 18\,\% of the time, with the majority being below ten minutes. To overcome this problem, recent literature has considered \ac{GNSS}-redundant solutions. \cite{heirich2017study, kroper2020using} find that railway tracks possess a unique ferromagnetic track signature that can be used for localization. However, changed track conditions or passing trains can disturb or change this signature. Other approaches have considered different kinds of signatures such as altitude \cite{heirich2017study}. In \cite{lavoie2021map}, the authors developed a map-aided train navigation based on \ac{IMU} measurements for subway systems.

To meet the high safety requirements for train positioning solutions, there is a need for redundant positioning methods using complementing sensors. Therefore, in this paper, we develop a train-borne \ac{PF} based positioning solution that can complement \ac{GNSS}, odometer, or magnetometer-reliant methods. It can also be used as an extension to improve filter performance in areas where \ac{GNSS} is sparsely available. \Cref{fig:introductionscheme} visualizes the examined situation. The key strength of the algorithm is minimalist sensor usage and an innovative formulation and usage of a \ac{DTM}. Only measurements of tangential acceleration, centripetal acceleration, and yaw angle turn rate - all recorded by one \ac{IMU} - and the railway track coordinates are required. We provide a flexible and adaptable formulation of the \ac{DTM}, which is incorporated in a \ac{PF} based positioning solution. The positioning problem is reduced to one dimension, leading to a two-dimensional state vector. Therefore, a low number of particles can be used, and computational costs are lower than in other \ac{PF} based solutions. The method is tested on the open data set \cite{WinterRailDataSetOctober2018}, which consists of a secondary line in a challenging environment.

%Terrain-Aided navigation:
%For exploiting track constraints, inspiration can be taken from terrain-aided navigation \cite{ma2023review} review on terrain-aided navigation for underwater vehicles

%Fancy methods:
%\cite{ko2022high} High-Speed Train Positioning Using Deep Kalman Filter With 5G NR Signals

\section{Enhanced geometric track map and track constraint formulation}
\label{sec:dtm}
\begin{figure} [h!]
    \centering
    \includegraphics[width=0.7\linewidth]{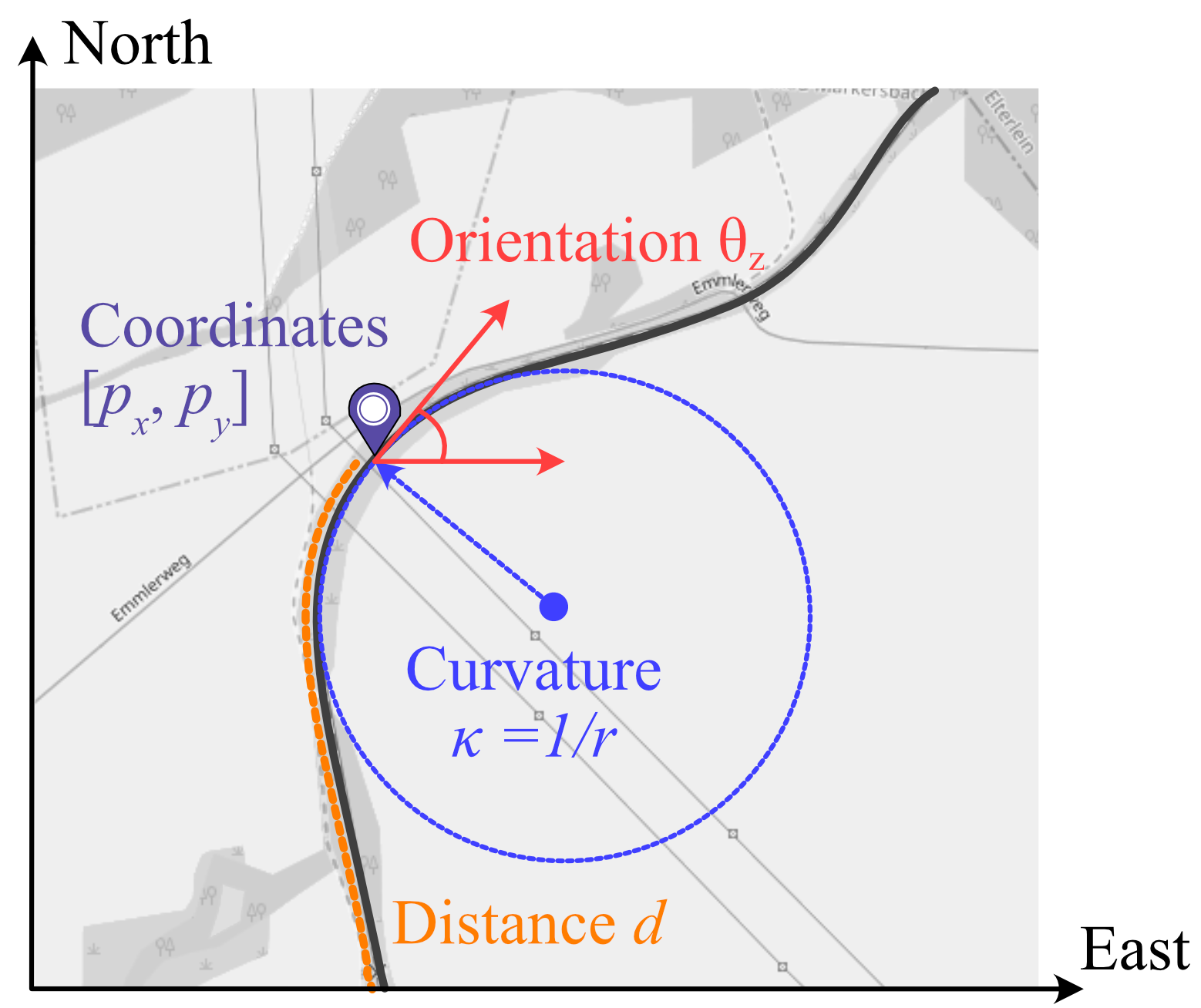}
    \caption{Visualization of variables stored from the track map}
    \label{fig:mapstructure_visual}
\end{figure}

The following section describes the design and usage of the \ac*{DTM} used in the estimation. The dynamic motion model explained in \Cref{sec:motionmodel} computes the train's position in 1D, the dimension being the traveled distance along the track. To acquire the corresponding map features and to map the position into 2D space, we require a geometric track map, which, in our proposed method, serves as a \ac{LUT}. Consider a set of $M$ discrete points along the track. The $i$th point along the track consists of an identifier $d_i$ and six features $\mu_i$ recorded at this point. The track map $ \mathcal{M} $ is expressed as follows:
\begin{align}
    \label{eqn:dtm}
    \mathcal{M} &= \left\{ \left(
        d_i, \boldsymbol{\mu}_i  \right) \right\} \\
    \boldsymbol{\mu}_i &= \begin{bmatrix}
        p_{x,i} & p_{y,i} & p_{z,i} & \kappa_i & \theta_{x,i} & \theta_{y,i} & \theta_{z,i}
    \end{bmatrix}
\end{align}
$d_i$ is an identifier of a recorded point along the track corresponding to the distance along the track. Distance $d_i = 0$ is set arbitrarily, e.g., at the starting station or the beginning of the map. Note that $d_i-d_{i-1}$ must be the distance along the track, not the Euclidean distance, to avoid underestimating the traveled distance.
The map features included are the following:
\begin{itemize}
    \item $p_{x,i}$, $p_{y,i}$ and $p_{z,i}$ correspond to the 3D geographical position of the point in a fixed reference frame.
    \item $\kappa_i$ is the computed curvature of the track, see \ref{sec:curvaturederiv}.
    \item $\theta_{x,i}$, $\theta_{y,i}$ and $\theta_{z,i}$ are the track orientation expressed in roll, pitch and yaw angle, respectively.
\end{itemize}
\Cref{fig:mapstructure_visual} gives a visual overview of the variables stored in the track map.
The orientation information is used for \ac{IMU} pre-processing in the filter, curvature is the main feature for tracking, and position is the filter output. Further features for filter extensions can be included in $\mathcal{M}$, such as details about stations, turnouts, or further track signatures.

\subsection{Derivation of map curvature} 
\label{sec:curvaturederiv} To calculate the track curvature, a twice differentiable parametric representation of the 2D plane curve is required:
\begin{equation}
    \gamma(d) = (x(d),y(d))
\end{equation} 
In the \ac{DTM} as presented in (\ref{eqn:dtm}), an arc-length parameterization on distance $d$ is given. A cubic spline interpolation between recorded map points ensures differentiability. The curvature is then estimated as
\begin{equation}
    \kappa(d) = \frac{x' y'' - y' x''}{(x'^2+y'^2)^{3/2}}
\end{equation}
% If arc-length parameterization is not given in an available data set, the method proposed in \cite{wang2002arc} may be applied.
Spline interpolation suffers from some drawbacks, e.g., closely spaced data points can lead to oscillations in the interpolated curve. In \cite{andersson2023compact}, the author found the most accurate computation of railway track curvature by reducing the data points used for spline interpolation using the Ramer-Douglas-Peucker algorithm and interpolating to the original density after spline interpolation. Importantly, this strategy does not alter the 2D coordinates stored in the \ac{LUT}.

\subsection{Map Matching and interpolation of map data}
\label{sec:interpolation}
To derive suitable parameters from the \ac{DTM}, we find the largest $d_i^{(\mathcal{M})}$ that is smaller than the estimated distance $d_k$ at time step $k$:
\begin{align}
\label{eqn:mapping1}
    q_1^{(\mathcal{M})} &= \mathrm{max}\{ d_i^{(\mathcal{M})} | d_i^{(\mathcal{M})} \in \mathcal{M} \, \mathrm{and} \, d_i^{(\mathcal{M})}<d_k \} \\
    q_2^{(\mathcal{M})} &= q_1^{(\mathcal{M})}+1
\end{align}
The parameter set $\mu_k$ retrieved from the \ac{DTM} is obtained by linear interpolation between $q_1^{(\mathcal{M})}$ and $q_2^{(\mathcal{M})}$:
\begin{align}
    r_k &= \frac{d_k - q_1^{(\mathcal{M})}}{|q_2^{(\mathcal{M})}-q_1^{(\mathcal{M})}|} \\
    \label{eqn:mapping2}
    \mu_k &= (1-r_k)\cdot \mu(q_1^{(\mathcal{M})})+r_k \cdot \mu (q_2^{(\mathcal{M})})
\end{align}
For the simplicity of the formulation, we assumed a forward travel direction.
Note that
\begin{equation}
    \left| \left| \begin{bmatrix}
        p_x & p_y
    \end{bmatrix}_{i+1}^{(\mathcal{M})} - \begin{bmatrix}
        p_x & p_y
    \end{bmatrix}_i^{(\mathcal{M})} \right| \right| \leq \left|d_{i+1}^{(\mathcal{M})} - d_i^{(\mathcal{M})}\right|
\end{equation}
We choose a linear interpolation between two recorded map points to compute $\mu_k$, which in a setup with 2D states would lead to underestimation of the traveled distance and velocity. Because of the format of the \ac{LUT}, this problem is eliminated in our approach. Further, other interpolation methods can easily be implemented if required. 

\section{Map-constrained particle filter}
\label{sec:method}
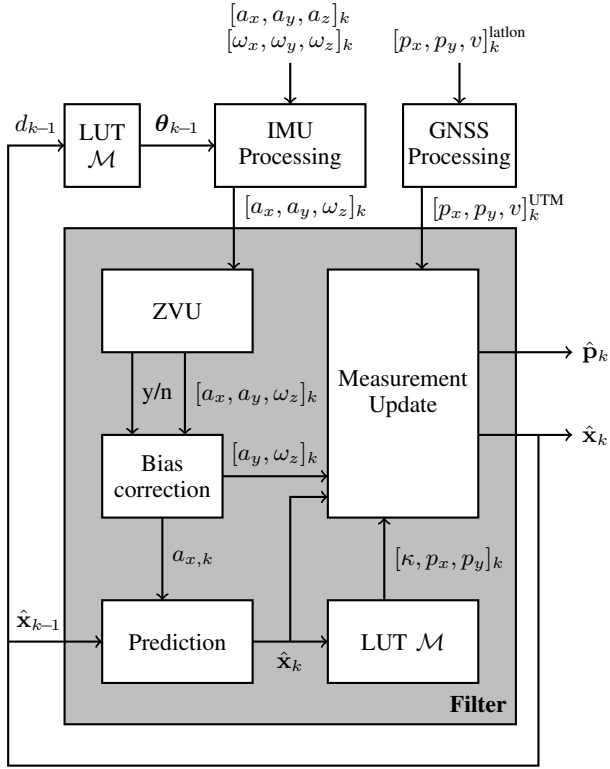
\begin{figure}[t]
    \centering
    \begin{tikzpicture}[xscale=1,yscale=1.1, every node/.style={font=\small}]
\draw[draw=black,thick,fill=lightgray] (0.5,0.5) rectangle (6.5,6.5);
\node[align=right] at (6,0.75) {\textbf{Filter}};

% Boxes inside PF
\draw[thick,draw=black,fill=white] (1,1) rectangle (3,2);
\node at (2,1.5) {Prediction};
\draw[thick,draw=black,fill=white] (1,3) rectangle (2.6,4) ;
\node[align=center] at (1.8,3.5) {Bias\\correction};
 \draw[thick,draw=black,fill=white] (1,5) rectangle (3,6);\node at (2,5.5) {ZVU};
\draw[thick,draw=black,fill=white] (4,1) rectangle (6,2);
\node at (5,1.5) {LUT $\mathcal{M}$};
\draw[thick,draw=black,fill=white] (4,3) rectangle (6,6);
\node[align=center] at (5,4.5) {Measurement\\Update};

% Vertical arrows inside the grey box
\draw[thick, ->] (1.8,3) -- (1.8,2) node[midway, right] {$a_{x,k}$};
\draw[thick, ->] (1.4,5) -- (1.4,4) node[midway, right] {y/n};
\draw[thick, ->] (2.1,5) -- (2.1,4) node[midway, right] {$[a_{x},a_y,\omega_z]_k$};
\draw[thick, ->] (4.75,2) -- (4.75,3) node[midway, right] {$[\kappa,p_x,p_y]_k$};
% horizontal arrows inside pd
\draw[thick, ->] (3,1.5) -- (4,1.5) node[midway, below] {$\hat{\mathbf{x}}_k$};
\draw[thick, ->] (2.6,3.5) -- (4,3.5) node[midway, above] {$[a_y,\omega_z]_k$};
%curved arrow
\draw[thick, ->] (3.5,1.5) -- (3.5,3.25) |- (4,3.25);

% Boxes outside pf
\draw[thick,draw=black,fill=white] (2.5,7) rectangle (4.5,8);
\node[align=center] at (3.5,7.5) {IMU\\Processing};
\draw[thick,draw=black,fill=white] (5,7) rectangle (6.5,8);
\node[align=center] at (5.75,7.5) {GNSS\\Processing};
\draw[thick,draw=black,fill=white] (0.5,7) rectangle (1.5,8);
\node[align=center] at (1,7.5) {LUT\\$\mathcal{M}$};

% Straight arrows outside
\draw[thick, ->] (3.5,8.5) node[align=center,above] {$[a_x,a_y,a_z]_k$\\ $[\omega_x,\omega_y,\omega_z]_k$} -- (3.5,8);
\draw[thick, ->] (5.75,8.5) node[above] {$[p_x,p_y,v]_k^{\text{latlon}}$} -- (5.75,8);
\draw[thick, ->] (2.75,7) -- (2.75,6) node[pos=0.25,right]{$[a_x,a_y,\omega_z]_k$};
\draw[thick, ->] (5.25,7) -- (5.25,6) node[pos=0.25,right]{$[p_x,p_y,v]_k^{\text{UTM}}$};
\draw[thick, ->] (1.5,7.5) -- (2.5,7.5) node[midway,above] {$\boldsymbol{\theta}_{k\!-\!1}$};
\draw[thick, ->] (6,4) -- (7.25,4) node[right] {$\hat{\mathbf{x}}_k$};
\draw[thick, ->] (6,5) -- (7.25,5) node[right] {$\hat{\mathbf{p}}_{k}$};
\draw[thick, ->] (-0.25,1.5)--(1,1.5) node[pos=0.33,above]{$\hat{\mathbf{x}}_{k\!-\!1}$};

% curved arrows
\draw[thick, ->] (6.8,4)--(6.8,0) |- (-0.25,0) |- (-0.25,7.5) |- (0.5,7.5) node[pos=0.75,above] {$d_{k\!-\!1}$};

\end{tikzpicture}
    \caption{Overview of complete algorithm}
    \label{fig:blockdiag}
\end{figure}

The following section describes the algorithm developed in this paper. An overview is given in \Cref{fig:blockdiag}.
We use a \ac{PF} based solution with a 2D linear state and a highly non-linear measurement update. The train's motion is modeled using a 1D constant acceleration model included in the particle diffusion step.

\subsection{Dynamic Motion Model}
\label{sec:motionmodel}
The following section corresponds to the Prediction block in \Cref{fig:blockdiag}.
Due to the constrained motion of the train along the tracks, we adopt a 1D motion model, as detailed below. Train motion consists of slow changes in acceleration. Hence, the constant acceleration model, as found in \cite{li2003survey} is a suitable model of the train dynamics. The state $\mathbf{x}$ at iteration $k$ consists of two variables:
\begin{equation}
    \mathbf{x}_k = \begin{bmatrix}
        d_k & v_k
    \end{bmatrix}^T
\end{equation}
with $v_k$ being velocity and $d_k$ traveled distance.
The constant acceleration model describes the motion dynamics as
\begin{align}
\label{eqn:motionmodel}
    \begin{bmatrix}
        d \\ v
    \end{bmatrix}_{k+1} = \begin{bmatrix}
        1 &T \\ 0& 1
    \end{bmatrix} \cdot \begin{bmatrix}
        d \\ v
    \end{bmatrix}_k &+ \begin{bmatrix}
        T^2/2 \\ T
    \end{bmatrix} \cdot u_k + \nu_k, \\
    \nu_k &\sim \mathcal{N}(0,\sigma_u^2) \notag
\end{align}
where $T$ is the sampling time. The system input $u$ is acceleration along the direction of movement measured by the \ac{IMU} $a_x^{\mathrm{(IMU)}}$. The measurement is corrected using the estimated bias, as detailed in \Cref{sec:iumprocessing}.

\subsection{Observation Model}
\label{sec:observation}
This section corresponds to the Measurement Update block in \Cref{fig:blockdiag}.
Consider the following measurements: 
\begin{equation}
    \mathbf{z}_k = [
        \underbrace{ \begin{matrix} p_x^{(\text{GNSS})} & p_y^{(\text{GNSS})} & v^{(\text{GNSS})} \end{matrix} }_{\text{if available}} \quad \begin{matrix} \omega_z^{(\text{IMU})} &  a_y^{(\text{IMU})}
    \end{matrix}]_k
\end{equation}
where $a_y^{(\text{IMU})}$ and $\omega_z^{(\text{IMU})}$ are measurements of the lateral acceleration and angular rate of the yaw angle. Both \ac{IMU} measurements were corrected using the estimated bias, as explained in \Cref{sec:iumprocessing}.
\ac{GNSS} measurements are generally not available except for a few irregular time intervals, which mostly reduces the measurement vector to
\begin{equation}
    \mathbf{z}_k = \begin{bmatrix}
        \omega_z^{(\text{IMU})} &  a_y^{(\text{IMU})}
    \end{bmatrix}_k
\end{equation}
The predicted measurement $\hat{\mathbf{z}}_k$ is expressed as a function of the state $\hat{\mathbf{x}}_k$ and the \ac{DTM} $ \mathcal{M} $:
\begin{equation}
\label{eqn:observationmodel}
    \hat{\mathbf{z}}_k = h_k(\hat{\mathbf{x}}_k, \mathcal{M}(\hat{d}_k))
\end{equation}
The estimated distance $\hat{d}_k$ has two purposes: it is included in the state variable $\hat{\mathbf{x}}_k$, but also the input to the \ac{LUT} $\mathcal{M}$.
Predicted measurements are calculated in three different ways: 
\begin{itemize}
    \item velocity $v$ is directly captured in the state
    \item position $p_x$, $p_y$ is a feature stored in the \ac{DTM}
    \item angular rate $\omega_z$ and lateral acceleration $a_y$ are calculated from a combination of state variables and map features.
\end{itemize}

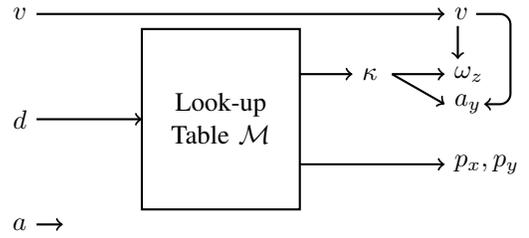
\begin{figure} [h!]
    \centering
    \begin{tikzpicture}[xscale=0.35,yscale=0.2]

    % Draw white-filled rectangle in the background
    \fill[white] (-2,-2.5) rectangle (19,14.5);
    
    % Draw the black box
    \draw[thick, ->] (0,13) node[left] {$v$} -- (15.5,13) node[right] {$v$};

    % Box
    \draw[thick, black] (4,0) rectangle (10,12);
    \node at (7,7) {Look-up};
    \node at (7,5) {Table $\mathcal{M}$};
    
    % Draw inputs
    \draw[thick, ->] (0,6) node[left] {$d$} -- (4,6);
    \draw[thick, ->] (0,-1) node[left] {$a$} -- (1,-1);

    % Outputs
    \draw[thick, ->] (10,9) -- (12,9) node[right] {$\kappa$};
    \draw[thick, ->] (13.5,9) -- (15.5,9) node[right] {$\omega_z$};
    \draw[thick, ->] (13.5,9) -- (15.5,7) node[right] {$a_y$};
    \draw[thick, ->] (10,3) -- (15.5,3) node[right] {$p_x,p_y$};

    % cross
    \draw[thick, ->] (16,12.2) -- (16,10);
    \draw[thick, ->, rounded corners] (16.7,13) -- (18,13) |- (17,7);
\end{tikzpicture}
    \caption{Visualization of observation model}
    \label{fig:observationmodel}
\end{figure}

 $h_k$ is specified in the following way:
 \begin{align}
 \label{eqn:virtualmeas}
     h_k ( \mathbf{x}_k,\mathcal{M} ) &= \begin{bmatrix}
         d_k \rightarrow \mathcal{M} \rightarrow p_{x,k} \\
         d_k \rightarrow \mathcal{M} \rightarrow p_{y,k} \\
         v_k \\
         v_k \cdot \kappa_k(d_k) \\
         v_k^2 \cdot \kappa_k(d_k)
     \end{bmatrix} + \mathbf{\eta}_k \\
     \kappa_k &= d_k \rightarrow \mathcal{M} \rightarrow \kappa_k
 \end{align}
Here, the right arrows indicate how $\mathcal{M}$ is used as a \ac{LUT}.
The relation in (\ref{eqn:virtualmeas}) is visualized in \Cref{fig:observationmodel}.

Note that it is not necessary to express the relation between the variables stored in $\mathcal{M}$ and distance $d_k$ in closed form.

\subsection{IMU Processing}
\label{sec:iumprocessing}
This section describes the \ac{IMU} processing, \ac{ZVU} and Bias Correction block shown in \Cref{fig:blockdiag}.

In the \ac{IMU} processing block, the distortion caused by gravitational acceleration is removed. This is achieved by transforming the 3D accelerometer data from the sensor frame to the inertial frame and removing the gravitational acceleration from $a_z^{\text{inertial}}$. Sensor frame and inertial frame are related via the orientation matrix $\mathbf{R}(\boldsymbol{\theta})$:
\begin{equation}
    \mathbf{a}^{\text{inertial}} = \mathbf{R}(\mathbf{\boldsymbol{\theta}}^{(\mathcal{M})}) \cdot \mathbf{a}^{\text{sensor}}
\end{equation}
where $\boldsymbol{\theta}$ is the orientation. In classic \ac{IMU} applications, orientation is estimated from the gyroscope data, which is sensitive to integration errors and bias itself. Therefore, we choose to compute $\boldsymbol{\theta}$ from the estimated distance of the previous filter update $\hat{d}_{k-1}$ using the \ac{DTM}, as described in \Cref{sec:interpolation}. The correction step is as follows:
\begin{equation}
\label{eqn:orientation}
    \mathbf{a}^{\text{sensor}} =\Tilde{\mathbf{a}}^{\text{sensor}} - \mathbf{R}^{-1}(\boldsymbol{\theta}) \begin{bmatrix}
        0 & 0 & g
    \end{bmatrix}^T
\end{equation}
where $\Tilde{\mathbf{a}}^{\text{sensor}}$ is the raw \ac{IMU} measurement. $\mathbf{a}^{\text{sensor}}$ is noted as $\mathbf{a}$ in the remaining paper.

Both accelerometer and gyroscope measurements contain several deterministic and stochastic error sources. The sensors are assumed to be calibrated in terms of scale factors, misalignment, and cross-coupling \cite{carlsson2021self}. The scaling error is assumed to be small in comparison to bias; therefore, the simplified error model is assumed as follows:
\begin{equation}
    \Tilde{\lambda} = \lambda + B + \delta B + \eta
\end{equation}
where $\lambda$ is the true value, $B$ is bias, $\delta B$ bias instability and $\eta$ the sensor noise. $\Tilde{\lambda}$ is the value measured by the accelerometer or gyroscope on one axis. 
To estimate $B$ and account for $\delta B$, $\Tilde{\lambda}$ is measured during a stand-still in the filter initialization. At each further stand-still, the bias estimate is updated using a moving average:
\begin{equation}
\label{eqn:biasestimation}
    B_k = 
    \begin{cases}
        \frac{1}{n}\left( \sum_{i=k-n+1}^{k-1} B_i +\Tilde{\lambda}_k \right) &\quad \text{stand-still} \\
        B_{k-1} &\quad \text{else}
    \end{cases}
\end{equation}
Each measurement is corrected by
\begin{equation}
\label{eqn:biasupdate}
    \lambda_k = \Tilde{\lambda}_k - B_k
\end{equation}
The paper refers to the estimated value after correction as sensor measurement.

To minimize integration errors and enable bias estimation updates, a Zero Velocity Update \ac{ZVU} is implemented. Stand-still is detected using a combined acceleration and angular rate detector \cite{callmer2010probabilistic}. If a stand-still is detected, $a_{x,k}^{(\text{IMU})}$, $a_{y,k}^{(\text{IMU})}$, $\omega_{z,k}^{(\text{IMU})}$, and $v_k$ are set to zero. 

\subsection{Complete Particle Filter Setup}
\label{sec:pfsetup}
In cases of linear state-space models, Kalman filters are proven to be the optimal filter. In this project, the motion model is linear; however, the observation model is highly non-linear. Therefore, we propose using a \ac{PF} to compute the position estimate using the motion model and observation model as described above. The complete algorithm is found in \Cref{alg:particle_filter}.

\begin{algorithm}[h!]
\caption{Particle filter with map matching}
\label{alg:particle_filter}
\begin{algorithmic}[1]
\State Initialize particles $\hat{\mathbf{x}}_0^{(j)} \sim p(\hat{\mathbf{x}}_0)$ and particle weights $ w_0^{(j)} = \frac{1}{N}  $ for $j = 1, \dots, N$ particles
\For{$k = 1, 2, \dots$}
\If{ZVU = 'true'}
\State Bias estimation update (\ref{eqn:biasestimation})
\State Set control input $u_k = 0$ 
\State Set measurements $a_{y,k} = \omega_{z,k} = 0$
\EndIf
\State Correct IMU measurements (\ref{eqn:orientation}), (\ref{eqn:biasupdate})
\For{$j = 1, \dots, N$}
\State Particle diffusion according to (\ref{eqn:motionmodel})
\State Map onto 2D plane using \ac{DTM}: (\ref{eqn:mapping1})-(\ref{eqn:mapping2})
\State Compute predicted measurements $\hat{\mathbf{z}}_k^{(j)}$ using (\ref{eqn:virtualmeas})
\State Compute weights: $w_k^{(j)} \propto p(\mathbf{z}_k|\hat{\mathbf{z}}_k^{(j)})$
\EndFor
\State Normalize weights: $\tilde{w}_k^{(j)} = \frac{w_k^{(j)}}{\sum_n w_k^{(n)}}$
\If{$N_{\mathrm{eff}}<N_{\mathrm{th}}$} \\{
    % Code to be executed if the condition is true
    Resample particles: sample $M_p$ particles with replacement from the current set of particles, using the weights $\tilde{w}_k^{(j)}$\;
  }
\EndIf
\State Compute $\hat{\mathbf{p}}_k$ (\ref{eqn:finalestimate})
\EndFor
\end{algorithmic}
\end{algorithm}
To avoid particle depletion, the particles are resampled when the effective sample size $N_{\mathrm{eff}}$ falls below a threshold $N_{\mathrm{th}} \in [1\, \, N]$. The effective sample size is commonly chosen as 
\begin{equation}
    N_{\mathrm{eff}} = \frac{1}{\sum_j\left( w_k^{(j)} \right)^2}
\end{equation}
Depending on the measurements, other resampling strategies may be sensible. If no absolute position measurements are available we suggest to resample when the train enters and travels on a curved track segment. 
The final position estimate is calculated using 
\begin{equation}
\label{eqn:finalestimate}
    \hat{\mathbf{p}}_k = \arg \  \underset{\mathbf{p}_k}{\max} \ p(\mathbf{p}_k|\mathbf{z}_{1:k},\mathbf{u}_{1:k})
\end{equation}
where $\hat{\mathbf{p}}_k = \begin{bmatrix}
    \hat{p}_{x,k} & \hat{p}_{y,k}
\end{bmatrix}^T$ is a sub-vector of $\hat{\mathbf{z}}_k$.

\section{Experimental Results}
\label{sec:results}
\begin{figure*}[t]
        \centering
        \input{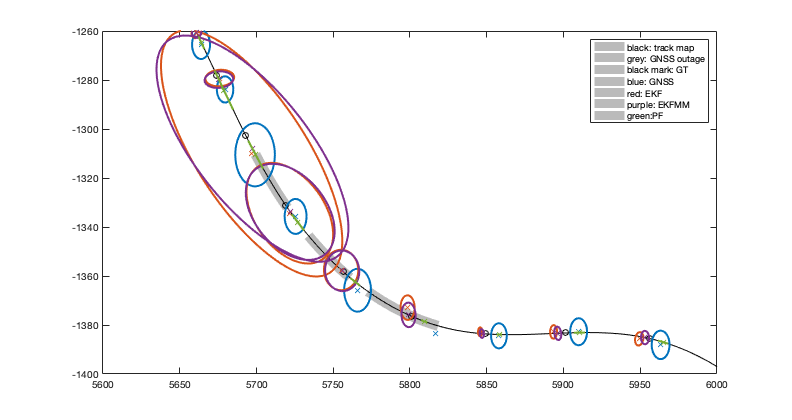}
        \caption{Comparison of different methods during a short GNSS outage}
        \label{fig:resultmap}
\end{figure*}
\begin{figure}[t]
    \centering
        \input{figs/maptotal.tex}
        \caption{Complete mapped test track}
        \label{fig:mapsatellite}
\end{figure}
\begin{figure}
    \centering
    \input{figs/velocityprofile}
    \caption{Velocity profile of the test drive}
    \label{fig:velocityprofile}
\end{figure}

For a proof-of-concept, we use the open Rail vehicle positioning data set, Lucy 2018 \cite{WinterRailDataSetOctober2018}. The data set contains 120\,km of conventional and secondary lines, including tight curves, steep slopes, and forested embankments. A 3D geometrical track map with centimeter-level precision is available on 24.2\,km of the track. The dataset contains 6 DOF \ac{IMU} measurements and position data from two \ac{GNSS} receivers. Since the positioning method presented in this paper heavily relies on the track map, we use the section of the dataset for which the 3D track map is available.

Since no ground truth is provided with the data set, the results in this paper are compared to the positions computed by the high-end \ac{GNSS} receiver by Thales mounted on the test train. To verify its accuracy, the across-track error of each measurement is calculated, showing that 95\% of all measurements lie within one meter from the track. Position updates are available every second with no outages during the test ride. To further enhance accuracy and establish what is referred to as ground truth throughout the remaining paper, the Thales \ac{GNSS} measurements are corrected by projection onto the track map. The other receiver is then used to test our proposed method.
\begin{figure*}[t]
    \centering
    \input{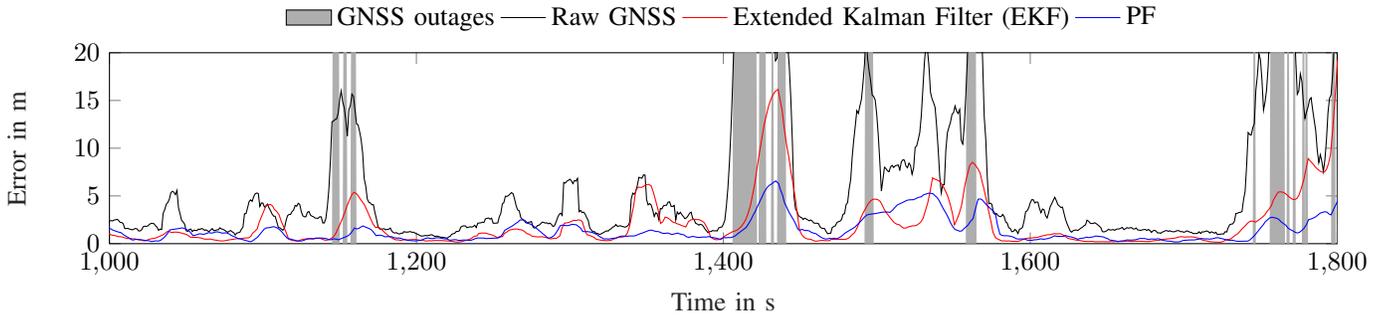}
    \caption{Absolute position error}
    \label{fig:abserrorlong}
\end{figure*}
\begin{figure}[h!]
    \centering
    \input{figs/cdf1.tex}
    \caption{Error \ac{CDF}}
    \label{fig:cdflong}
\end{figure}

All results are computed using our proposed method and compared to an \ac{EKF}. The used data set includes an \ac{EKF} based positioning solution, using tangential acceleration $a_x^{(\text{IMU})}$ and yaw angle turn rate $\omega_z^{(\text{IMU})}$ as input to a \ac{CTRA} motion model. \ac{GNSS} measurements of position and velocity are used in the measurement update. We combine this solution with the approach given in \cite{loffler2022using} to include map matching into the algorithm, which in the remaining paper will be called \ac{EKFMM}. The parameter settings are adjusted to the sensor specification and are listed for both methods in \Cref{tab:parameters}. The \ac{PF} uncertainties are increased compared to the \ac{EKFMM} to prevent particle depletion.
\begin{figure*}[t]
     \centering
     \begin{subfigure}[b]{0.48\textwidth}
         \centering
         \includegraphics[width=\textwidth]{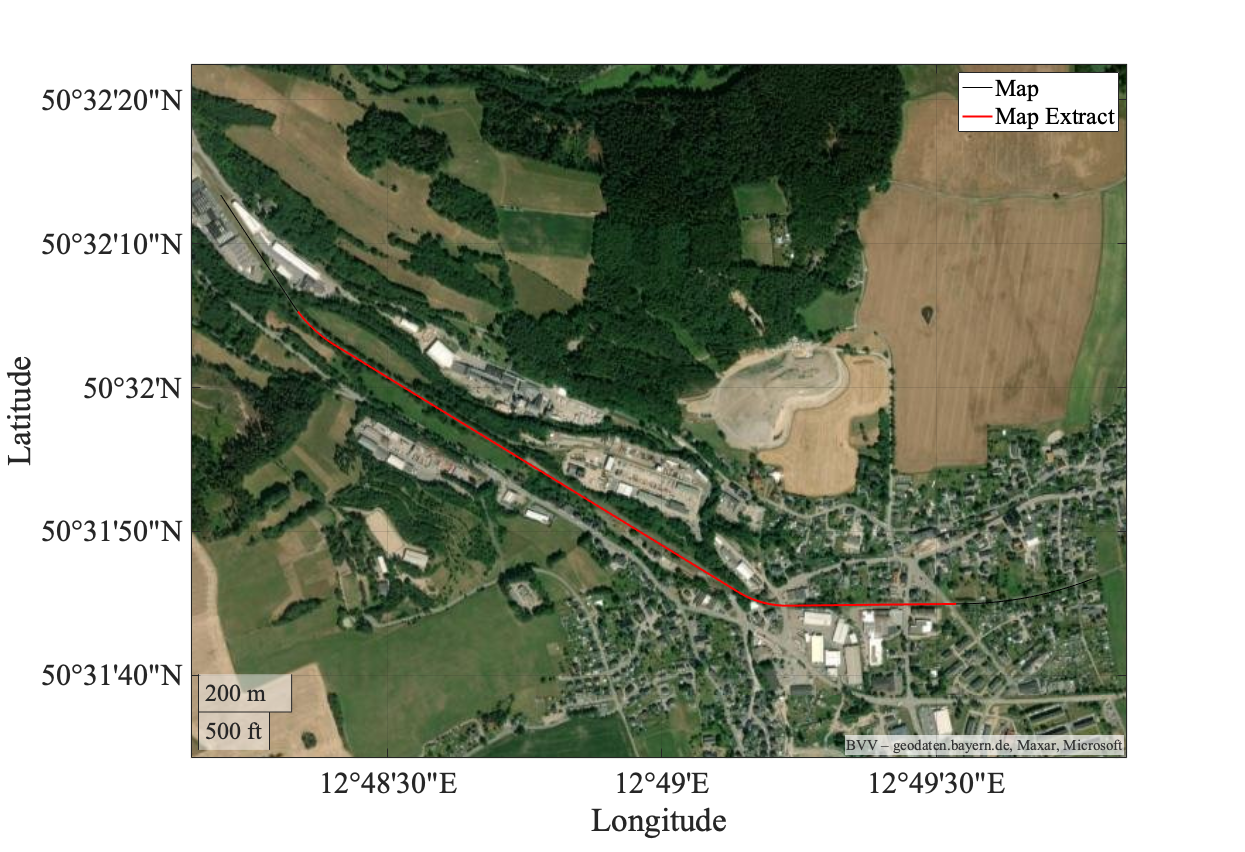}
         \caption{Straight passage from the track map}
         \label{fig:mapshortstraight}
     \end{subfigure}
     \hfill
     \begin{subfigure}[b]{0.48\textwidth}
         \centering
         \includegraphics[width=\textwidth]{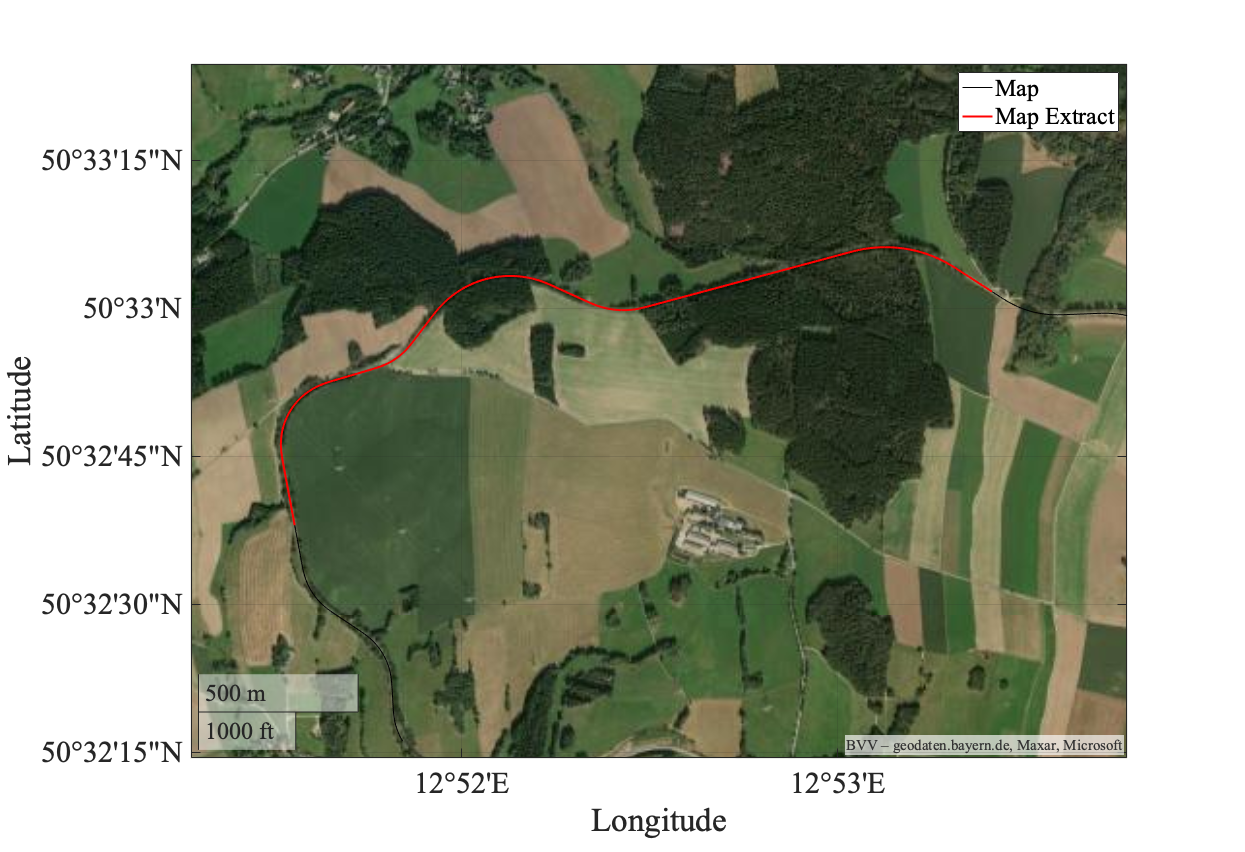}
         \caption{Curved passage of the track map}
         \label{fig:mapshortcurvy}
     \end{subfigure}
     \begin{subfigure}[b]{0.48\textwidth}
         \centering
         \input{figs/curv_straight}
         \caption{Curvature of straight track segment}
         \label{fig:curvstraight}
     \end{subfigure}
     \begin{subfigure}[b]{0.48\textwidth}
         \centering
         \input{figs/curv_curvy}
         \caption{Curvature of curved track segment}
         \label{fig:curvcurvy}
     \end{subfigure}
     \begin{subfigure}[b]{0.48\textwidth}
         \centering
         \input{figs/abserror_straight}
         \caption{Euclidean position error at straight track segment}
         \label{fig:abserrorstraight}
     \end{subfigure}
     \begin{subfigure}[b]{0.48\textwidth}
         \centering
         \input{figs/abserror_curvy}
         \caption{Euclidean position error at curved track segment}
         \label{fig:abserrorcurvy}
     \end{subfigure}
%     \begin{subfigure}[b]{0.48\textwidth}
%         \centering
%         \input{figs/cdf_straight}
%         \caption{\ac{CDF} of the $3\sigma$ positioning error of straight track segment}
%         \label{fig:cdfstraight}
%     \end{subfigure}
%     \begin{subfigure}[b]{0.48\textwidth}
%         \centering
%         \input{figs/cdf_curvy}
%         \caption{\ac{CDF} of the $3\sigma$ positioning error of curved track segment}
%         \label{fig:cdfcurvy}
%     \end{subfigure}
     \caption{Performance evaluation during indefinite \ac{GNSS} outage on a straight and on a curved path segment}
     \label{fig:bigplot}
\end{figure*}
% \usepackage{booktabs}

\begin{table}[h!]
\centering
\caption{Parameter values used in both filters}
\label{tab:parameters}
\begin{tabular}{ccccc}
\hline
\multicolumn{1}{c|}{} & $\sigma_{a_x}^{(\text{IMU})}$ & $\sigma_{a_y}^{(\text{IMU})}$ & $\sigma_{\omega_z}^{(\text{IMU})}$ & $\sigma_{\text{bias}}$ \\ \hline
\multicolumn{1}{c|}{\ac{PF}} & 0.01\,g & 0.01\,g & 0.2\,$^{\circ}$/s & 5e-6\,m/s$^2$ \\ 
\multicolumn{1}{c|}{\ac{EKFMM}} & 0.005\,g & - & 0.05\,$^{\circ}$/s & - \\ \hline
 &  &  &  &  \\ 
\hline
\multicolumn{1}{c|}{} & $\sigma_{p_x}^{(\text{GNSS})}$ & $\sigma_{p_y}^{(\text{GNSS})}$ & $\sigma_v^{(\text{GNSS})}$ & $\sigma^{(\text{map})}$ \\ \hline
\multicolumn{1}{c|}{\ac{PF}} & 2.45\,m* & 4.13\,m* & 0.4\,m/s* & - \\ 
\multicolumn{1}{c|}{\ac{EKFMM}} & 2.04\,m* & 3.45\,m* & 0.35\,m/s* & 0.01\,m \\ \hline 
\end{tabular} \\
\flushleft
* Uncertainty updated dynamically according to \ac{GNSS} receiver statistics. The displayed value is the mean uncertainty
\end{table}

Two scenarios are presented. First, the method is tested and evaluated on experimental data given in the data set. Second, the experimental data is altered by removing \ac{GNSS} data after 50\,s. In both scenarios, 1000 particles are used. The value was found empirically to balance computational cost and filter convergence.

\subsection{Evaluation using experimental data}
\label{sec:resrealdata}

In this section, the proposed method is tested on experimental data. The map of the test track is shown in \Cref{fig:mapsatellite}. 

The test train's starting position and travel direction are marked in \Cref{fig:resultmap}, and the velocity profile is shown in \Cref{fig:velocityprofile}. The recorded data includes stops, varying speeds, and several \ac{GNSS} outages along the journey, the longest being 35\,s. A representative example of the filter's performance can be seen in \Cref{fig:resultmap}, which shows position estimates and estimated $3\sigma$ confidence interval of the raw \ac{GNSS} data, \ac{EKFMM} solution and the proposed \ac{PF} based solution, predicted according to the respective model. The confidence intervals of the \ac{EKFMM} solution quickly increase during the outage, while the \ac{PF} solution can maintain the position estimate with stable uncertainty. 
The error statistics are shown in \Cref{fig:abserrorlong} and \Cref{fig:cdflong}. Both the \ac{EKFMM} and the \ac{PF} perform similarly when \ac{GNSS} is availble, but during \ac{GNSS} outages marked in grey, the \ac{PF} solution outperforms the \ac{EKFMM}. For the \ac{PF} solution, the $3\sigma$ error lies at 11.3\,m, whereas for the \ac{EKFMM} solution, the $3\sigma$ error is 17.71\,m.

\subsection{Indefinite \ac{GNSS} outage on different track characteristics}
\label{sec:resLongOutage}

This section examines the performance of the \ac{PF} solution during an indefinite \ac{GNSS} outage. The experimental data as in \Cref{sec:resrealdata} is used, but the \ac{GNSS} signal is removed after 50\,s, simulating an indefinite outage. The performance is evaluated on two sections of the test track with different track characteristics. \Cref{fig:mapshortstraight} shows a segment consisting of several straight sections. A plot of the track curvature is shown in \Cref{fig:curvstraight}. In contrast, \Cref{fig:mapshortcurvy} shows a segment consisting of several turns and short straight sections. The corresponding curvature is given in \Cref{fig:curvcurvy}. \ac{IMU} measurements and track coordinates are as provided by the dataset, but \ac{GNSS} reception stops after 50\,s. We compare the results using the proposed method to the \ac{EKFMM} by computing the absolute position error, see \Cref{fig:abserrorstraight,fig:abserrorcurvy}. It can be seen that the \ac{PF} solution performance is dependent on the track characteristic. The absolute position error increases on straight track sections. During curves, the accuracy reached by the proposed method is comparable to the accuracy when \ac{GNSS} is available. During straight sections, both methods need to rely on the accelerometer measurements and diverge. However, it can be seen that the \ac{PF}, in contrast to the \ac{EKFMM} can recover to high accuracy during short curves. In \Cref{fig:abserrorstraight} after 190\,s, the absolute error does not increase further. This is due to four shorter stand-still phases and the resulting sensor bias update.
%An interesting detail is that the average error decreases shortly before a curve and then increases again. This can be explained by an overcompensation: the estimate relies on the tangential acceleration measurement on a straight track segment, which is biased. Therefore, the forecast of both position and velocity experience and increasing bias during a straight segment. Shortly before a curve, all particles entering the curve are eliminated, leading to a decrease of the estimated velocity, and hence, a short-term increase of the position error. 

\section{Conclusion}
\label{sec:conclusion}
In this paper, we have proposed a minimalist approach for train localization during \ac{GNSS} outages by utilizing track geometry and \ac{IMU} sensor data. A discrete track map has been formulated as a look-up table and used in different ways: to map the 1D position estimate of the \ac{PF} onto the 2D plane. Also, geometry features such as curvature have been used in the measurement update with gyroscope measurements. By this, the position estimation in the \ac{PF} has been reduced to one dimension, resulting in lower computational complexity of the \ac{PF} compared to 2D or 3D tracking filters. The formulation as a \ac{LUT} ensures great flexibility and adaptability of the approach. Map coordinates do not need to be placed equidistantly. Furthermore, any interpolation method between map coordinates can be implemented and computed offline. Additional features and measurements can be added easily. 

The proposed method was tested on experimental data from the open Lucy 2018 data set. The results show that the \ac{PF} outperforms the \ac{EKFMM} approach during shorter \ac{GNSS} outages and providing similar accuracy when \ac{GNSS} is available. We simulated an indefinite \ac{GNSS} outage and tested the filter performance on two segments of the test track with different characteristics. It could be shown that the filter can provide an accurate positioning solution during track curves and curvy railway lines. However, the performance is dependent on the track characteristic and does, therefore, not work as a stand-alone positioning solution, when only using the \ac{IMU}.

The key strength of the approach lies in its minimalist sensor usage and flexible track map incorporation. To meet the high safety standards for train positioning solutions, redundant solutions must be implemented. The presented approach does not use sensors incorporated in recent train positioning solutions, such as odometer data or magnetometers. If knowledge of the starting position is available, even the irregular \ac{GNSS} measurements can be removed. Therefore, the proposed method is practical to include as redundant positioning method, increasing safety of the positioning solution. Another possible application is an implementation in a classical \ac{GNSS}/\ac{IMU} sensor fusion approach to increase positioning accuracy in challenging environments. Further research will include testing in different scenarios and an improved bias correction.

\bibliographystyle{ieeetr}
\bibliography{references}

\begin{thebibliography}{10}

\bibitem{hamid2020impact}
H.~A. Hamid, G.~L. Nicholson, and C.~Roberts, ``Impact of train positioning inaccuracies on railway traffic management systems: framework development and impacts on tms functions,'' {\em IET Intelligent Transport Systems}, vol.~14, no.~6, pp.~534--544, 2020.

\bibitem{gustafsson2010statistical}
F.~Gustafsson, {\em Statistical {S}ensor {F}usion}.
\newblock Studentlitteratur, 2010.

\bibitem{en50126}
``{EN50126} - {R}ailway applications - {T}he specification and demonstration of {R}eliability, {A}vailability, {M}aintainability, and {S}afety ({RAMS}),'' {\em Eur. Commitee for Electrotechnical Standardization Std.}

\bibitem{reimer2016ins}
C.~Reimer, F.~M{\"u}ller, and E.~Hin{\"u}ber, ``{INS}/{GNSS}/odometer data fusion in railway applications,'' in {\em 2016 DGON Intertial Sensors and Systems (ISS)}, pp.~1--14, IEEE, 2016.

\bibitem{otegui2018evaluation}
J.~Otegui, A.~Bahillo, I.~Lopetegi, and L.~E. D{\'\i}ez, ``Evaluation of experimental {GNSS} and 10-{DOF} {MEMS} {IMU} measurements for train positioning,'' {\em IEEE Transactions on Instrumentation and Measurement}, vol.~68, no.~1, pp.~269--279, 2018.

\bibitem{heirich2016bayesian}
O.~Heirich, ``Bayesian train localization with particle filter, loosely coupled {GNSS}, {IMU}, and a track map,'' {\em Journal of Sensors}, vol.~2016, 2016.

\bibitem{liu2016track}
J.~Liu, B.~Cai, and J.~Wang, ``Track-constrained {GNSS}/odometer-based train localization using a particle filter,'' in {\em 2016 IEEE Intelligent Vehicles Symposium (IV)}, pp.~877--882, IEEE, 2016.

\bibitem{liu2023robust}
D.~Liu, W.~Jiang, B.~Cai, O.~Heirich, J.~Wang, and W.~Shangguan, ``Robust train localisation method based on advanced map matching measurement-augmented tightly-coupled {GNSS}/{INS} with error-state {UKF},'' {\em The Journal of Navigation}, pp.~1--24, 2023.

\bibitem{winter2018increasing}
H.~Winter, V.~Willert, and J.~Adamy, ``Increasing accuracy in train localization exploiting track-geometry constraints,'' in {\em 2018 21st International Conference on Intelligent Transportation Systems (ITSC)}, pp.~1572--1579, IEEE, 2018.

\bibitem{heirich2013railslam}
O.~Heirich, P.~Robertson, and T.~Strang, ``Rail{SLAM}-localization of rail vehicles and mapping of geometric railway tracks,'' in {\em 2013 IEEE International Conference on Robotics and Automation}, pp.~5212--5219, IEEE, 2013.

\bibitem{loffler2022using}
W.~L{\"o}ffler and M.~Bengtsson, ``Using probabilistic geometrical map information for train localization,'' in {\em 2022 25th International Conference on Information Fusion (FUSION)}, pp.~01--08, IEEE, 2022.

\bibitem{loffler2022evaluating}
W.~L{\"o}ffler and M.~Bengtsson, ``Evaluating the impact of map inaccuracies on path discrimination behind railway turnouts,'' in {\em 2022 IEEE 95th Vehicular Technology Conference: (VTC2022-Spring)}, pp.~1--5, IEEE, 2022.

\bibitem{hasberg2011simultaneous}
C.~Hasberg, S.~Hensel, and C.~Stiller, ``Simultaneous localization and mapping for path-constrained motion,'' {\em IEEE Transactions on Intelligent Transportation Systems}, vol.~13, no.~2, pp.~541--552, 2011.

\bibitem{winter2019generating}
H.~Winter, S.~Luthardt, V.~Willert, and J.~Adamy, ``Generating compact geometric track-maps for train positioning applications,'' in {\em 2019 IEEE Intelligent Vehicles Symposium (IV)}, pp.~1027--1032, IEEE, 2019.

\bibitem{andersson2023compact}
A.~Andersson, ``Compact digital track maps: Enhancing train traveller information at the crossing of accuracy and availability: A comparative analysis of algorithms for generating compact representations of railway tracks,'' No.~{TRITA}-{EECS}-{EX}-2023:712, (KTH Royal Institute of Technology), 2023.

\bibitem{spinsante2020hybridized}
S.~Spinsante and C.~Stallo, ``Hybridized-{GNSS} approaches to train positioning: {C}hallenges and open issues on uncertainty,'' {\em Sensors}, vol.~20, no.~7, p.~1885, 2020.

\bibitem{zhu2018gnss}
N.~Zhu, J.~Marais, D.~B{\'e}taille, and M.~Berbineau, ``{GNSS} position integrity in urban environments: A review of literature,'' {\em IEEE Transactions on Intelligent Transportation Systems}, vol.~19, no.~9, pp.~2762--2778, 2018.

\bibitem{d2023innovative}
P.~d’Harcourt, J.-P. Michel, L.~Poletti, and S.~Glevarec, ``Innovative and efficient inertial navigation system for train localization in {GNSS}-denied environments,'' in {\em 2023 DGON Inertial Sensors and Systems (ISS)}, pp.~1--18, IEEE, 2023.

\bibitem{heirich2017study}
O.~Heirich, B.~Siebler, and E.~Hedberg, ``Study of train-side passive magnetic measurements with applications to train localization,'' {\em Journal of Sensors}, vol.~2017, 2017.

\bibitem{kroper2020using}
B.~Kr{\"o}per, M.~Lauer, and M.~Spindler, ``Using the ferromagnetic fingerprint of rails for velocity estimation and absolute localization of railway vehicles,'' in {\em 2020 European Navigation Conference (ENC)}, pp.~1--10, IEEE, 2020.

\bibitem{lavoie2021map}
M.-A. Lavoie and J.~R. Forbes, ``Map-aided train navigation with {IMU} measurements,'' in {\em 2021 IEEE/RSJ International Conference on Intelligent Robots and Systems (IROS)}, pp.~3465--3470, IEEE, 2021.

\bibitem{WinterRailDataSetOctober2018}
H.~Winter, ``Rail vehicle positioning data set: Lucy, october 2018,'' 2020.

\bibitem{li2003survey}
X.~R. Li and V.~P. Jilkov, ``Survey of maneuvering target tracking. {P}art {I}. {D}ynamic models,'' {\em IEEE Transactions on aerospace and electronic systems}, vol.~39, no.~4, pp.~1333--1364, 2003.

\bibitem{carlsson2021self}
H.~Carlsson, I.~Skog, and J.~Jald{\'e}n, ``Self-calibration of inertial sensor arrays,'' {\em IEEE Sensors Journal}, vol.~21, no.~6, pp.~8451--8463, 2021.

\bibitem{callmer2010probabilistic}
J.~Callmer, D.~T{\"o}rnqvist, and F.~Gustafsson, ``Probabilistic stand still detection using foot mounted {IMU},'' in {\em 2010 13th International Conference on Information Fusion}, pp.~1--7, IEEE, 2010.

\end{thebibliography}

\end{document}